\documentclass[dvipsnames]{article} %
\usepackage{colm2024_conference}

\usepackage{booktabs}
\usepackage{enumitem}
\usepackage{wrapfig}
\usepackage{algorithm}
\usepackage{algpseudocode}
\usepackage{graphicx}
\usepackage{subfigure}
\usepackage[misc]{ifsym}

\usepackage{microtype}
\usepackage{amsmath}
\usepackage{colortbl}
\usepackage[utf8]{inputenc}
\usepackage[T1]{fontenc}
\definecolor{lightgray}{rgb}{0.9,0.9,0.9}
\usepackage{caption}
\usepackage{subcaption}
\usepackage{setspace}
\usepackage{url}
\usepackage{multirow}
\usepackage{tabularx}
\usepackage{blindtext}
\usepackage{pgfplots}
\pgfplotsset{compat=1.18} 
\usepackage{tikz}
\usetikzlibrary{er,positioning,bayesnet}
\usepackage{makecell}
\usepackage{tipa}
\usepackage{siunitx}
\usepackage{nicefrac}
\usepackage{listings}
\usepackage[raster,skins, most]{tcolorbox} %
\usepackage{xltabular}
\usepackage{adjustbox}
\usepackage{xurl}
\usepackage{rotating}
\usepackage[normalem]{ulem}

\usepackage{fontawesome}

\useunder{\uline}{\ul}{}

\usepackage{amsmath,amsfonts,bm}

\def\eqref#1{equation~\ref{#1}}

\def\1{\bm{1}}

\DeclareMathAlphabet{\mathsfit}{\encodingdefault}{\sfdefault}{m}{sl}
\SetMathAlphabet{\mathsfit}{bold}{\encodingdefault}{\sfdefault}{bx}{n}

\renewcommand{\ghlink}{https://github.com/Alibaba-NLP/DeepResearch}

\usepackage{makecell}
\usetikzlibrary{tikzmark}
\makeatletter
\newcommand*\myfontsize{%
  \@setfontsize\myfontsize{7}{8}%
}
\makeatother

\usepackage{fontawesome}   

\definecolor{uclablue}{RGB}{159, 195, 224}

\definecolor{uclagold}{RGB}{255, 240, 180}

\definecolor{aliceblue}{RGB}{255, 238, 241}

\definecolor{cadmiumgreen}{rgb}{0.0, 0.42, 0.24}

\definecolor{myred}{rgb}{0.7, 0.3, 0.0}
\definecolor{myblue}{rgb}{0.2, 0.3, 0.6}
\definecolor{babygreen}{rgb}{0.85, 0.97, 0.85}

\definecolor{purple1}{RGB}{126, 107, 196}
\definecolor{purple2}{RGB}{199, 158, 207}
\definecolor{purple3}{RGB}{214, 200, 255}
\definecolor{purple4}{RGB}{254, 240, 255}

\definecolor{deepblue}{RGB}{48, 58, 82}

\makeatletter

\newcommand{\symboletongyi}{\raisebox{0pt}{~\includegraphics[scale=0.012]{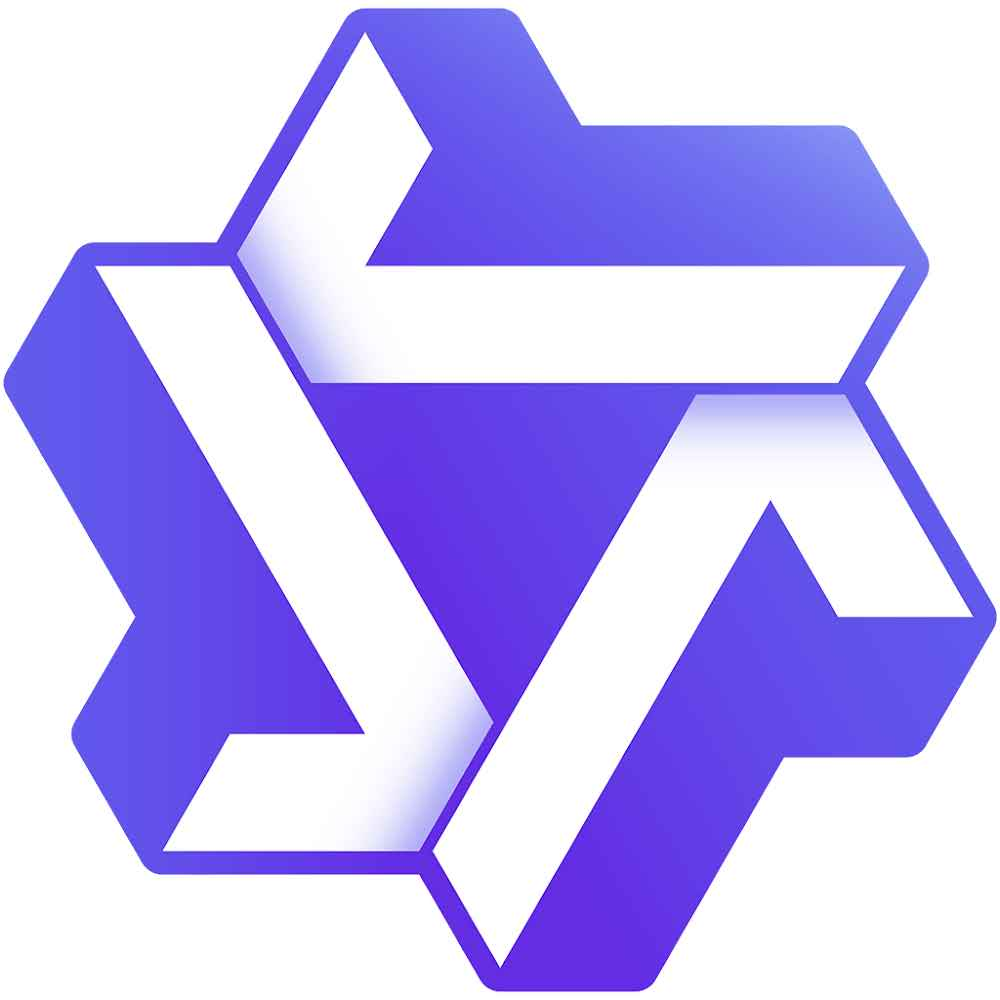}}~}

\definecolor{deepPurple}{HTML}{330066}

\definecolor{uclablue_old}{rgb}{0.15, 0.45, 0.68}
\hypersetup{
    breaklinks,
    citecolor=uclablue_old,
    colorlinks=true,
}

\newtcolorbox{mybox}[2][]
  {colback = black!5!white, colframe = black!75!black, fonttitle = \bfseries,
    colbacktitle = black!100!black, enhanced, before upper={\fontsize{8}{11}\obeyspaces\obeylines\selectfont}, fontupper=\selectfont,
    attach boxed title to top left={yshift=-2.2mm,xshift=4mm},
    title=#2,#1}

\newcommand{\equal}{\textsuperscript{\dag}}

\definecolor{darkgreen}{rgb}{0.0, 0.5, 0.0}

\author{%
\small{Zhaopeng Feng\equal$^{(\textrm{\Letter})}$, Liangcai Su\equal, Zhen Zhang\equal, Xinyu Wang$^{(\textrm{\Letter})}$, Xiaotian Zhang \\
Xiaobin Wang, Runnan Fang, Qi Zhang, Baixuan Li, Shihao Cai, Rui Ye, Hui Chen, Jiang Yong \\ 
Joey Tianyi Zhou, Chenxiong Qian, Pengjun Xie, Bryan Hooi, Zuozhu Liu, Jingren Zhou}%
  \\[1em]               
  {\fontsize{10pt}{11pt}\selectfont          
Tongyi Lab\symboletongyi, Alibaba Group}\\
}

\usepackage{caption}
\usepackage{subcaption}
\usepackage{amssymb}

\definecolor{mydarkgray}{gray}{0.2} 

\begin{document}

\title{%
\hspace{-2.5em} 
\raisebox{-2.17em}{
  \parbox[t]{0.2in}{\includegraphics[width=0.65in]{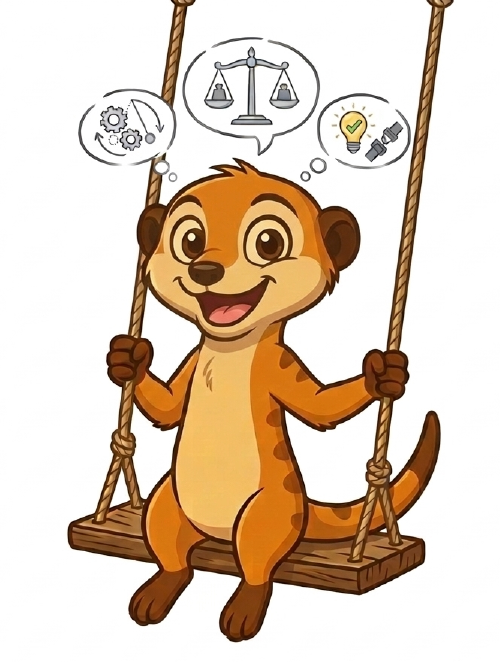}} 
  \hspace{1.3em}
}
\begin{tabular}[t]{l} 
  \parbox[t]{0.8\textwidth}{\centering 
    \textsc{AgentSwing}: Adaptive Parallel Context  Management \\ Routing for Long-Horizon Web Agents
  }
\end{tabular}
}
\maketitle

\begingroup
\renewcommand\thefootnote{\equal}
\footnotetext{Equal contribution.}
\endgroup

\begingroup
\renewcommand\thefootnote{$^{\textrm{\Letter}}$}
\footnotetext{Correspondence to: \{\texttt{zhaopengfeng424}, \texttt{wangxinyu.nlp\}@gmail.com}.}
\endgroup

\begin{abstract}
  As large language models (LLMs) evolve into autonomous agents for long-horizon information-seeking, managing finite context capacity has become a critical bottleneck. Existing context management methods typically commit to a single fixed strategy throughout the entire trajectory. Such static designs may work well in some states, but they cannot adapt as the usefulness and reliability of the accumulated context evolve during long-horizon search. To formalize this challenge, we introduce a probabilistic framework that characterizes long-horizon success through two complementary dimensions: \textit{\textbf{search efficiency}} and \textit{\textbf{terminal precision}}. Building on this perspective, we propose \textbf{AgentSwing}, a state-aware adaptive parallel context management routing framework. At each trigger point, AgentSwing expands multiple context-managed branches in parallel and uses lookahead routing to select the most promising continuation. Experiments across diverse benchmarks and agent backbones show that AgentSwing consistently outperforms strong static context management methods, often matching or exceeding their performance with up to $3\times$ fewer interaction turns while also improving the ultimate performance ceiling of long-horizon web agents. Beyond the empirical gains, the proposed probabilistic framework provides a principled lens for analyzing and designing future context management strategies for long-horizon agents.
\end{abstract}

\section{Introduction}

As large language models (LLMs) evolve from single-turn question answering assistants into autonomous agents capable of web browsing and sequential tool use, long-horizon information-seeking has emerged as a critical testbed of their real-world capabilities~\citep{webwalker,wu2025webdancerautonomousinformationseeking,kimi_researcher,fang2025towards,li2025websailornavigatingsuperhumanreasoning,tao2025webshaper,li2025websailorv2bridgingchasmproprietary}. In such tasks, solving a problem often requires tens or even hundreds of steps of searching, visiting, verifying, and backtracking before the agent can locate the key evidence and produce a final answer.


A central bottleneck in deep information-seeking is the tension between finite context capacity and the need for long-horizon exploration~\citep{bc_en,hle,wong2025widesearch}. 
Under a fixed context budget, an agent may exhaust its workspace before completing a sufficiently informative search trajectory. As a result, \textbf{context management} has become a key mechanism shaping the performance ceiling of long-horizon agents~\citep{context-engineering,liu2025deepseek}. Recent frontier systems have shown that aggressive context management, such as \textit{Discard-All}, can substantially improve long-horizon performance by enabling agents to discard accumulated context to sustain more interaction turns~\citep{liu2025deepseek,team2026kimik2.5,zeng2026glm}. Most existing context management approaches rely on a single fixed strategy that is repeatedly applied throughout the entire trajectory. This design is inherently limited in long-horizon search, where the quality of the accumulated context evolves over time. Some trajectory states contain useful intermediate structures that should be retained, while others are dominated by noise, drift, or unproductive search history and therefore call for more aggressive intervention.

To make this limitation explicit, we introduce the first probabilistic perspective for deep information-seeking agents that characterizes success through two complementary dimensions: \textbf{search efficiency} and \textbf{terminal precision}. Search efficiency measures whether an agent can reach a stopping point before exhausting available resources, while terminal precision measures whether the final answer is correct conditioned on reaching such a stopping point. 
This view reveals that commonly reported metrics such as Pass@1 or accuracy are not monolithic indicators in long-horizon settings. 
Instead, end-to-end success depends jointly on whether the agent can arrive at a terminal state with the final answer and whether it can answer correctly once there.

Building on this perspective, we propose \textbf{AgentSwing}, an adaptive parallel context management routing framework for long-horizon web agents.
Instead of committing to a single context management operation at every trigger point, AgentSwing expands multiple context-managed branches from the current trajectory state and uses a lookahead routing mechanism to select the most promising continuation.  
In this way, AgentSwing leverages the complementary strengths of heterogeneous context management strategies and moves beyond the efficiency-precision trade-off of static context management methods. Experiments on \begin{wrapfigure}{r}{7.5cm}
    \centering
    \vspace{-1em}
    \includegraphics[width=\linewidth]{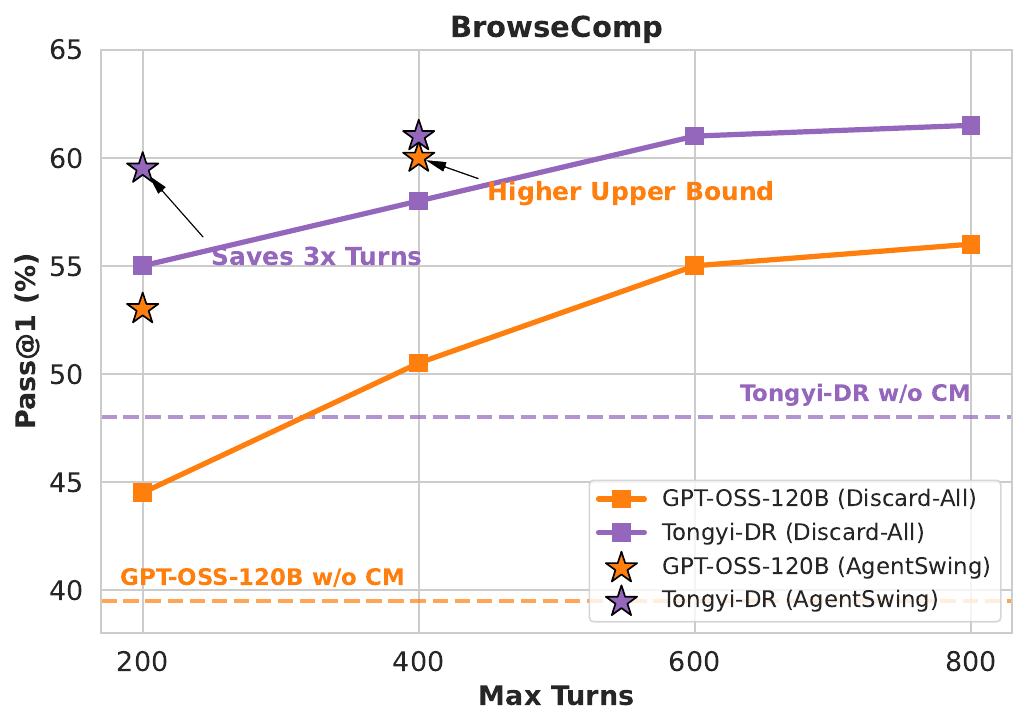}
    \caption{Performance on BrowseComp under different interaction budgets. Dashed lines denote the baselines without context management.}
    \vspace{-1.5em} 
    \label{fig:intro_turns}
\end{wrapfigure}several challenging long-horizon benchmarks with diverse open-source backbones, including GPT-OSS-120B~\citep{gptoss}, DeepSeek-v3.2~\citep{liu2025deepseek}, and Tongyi-DR-30B-A3B~\citep{tongyidr}, show that AgentSwing consistently outperforms strong static methods. Under constrained interaction budgets, it reaches or exceeds the performance of static strategies that require up to \(3\times\) more interaction turns, while also achieving a higher ultimate performance ceiling (see Figure~\ref{fig:intro_turns}). It pushes DeepSeek-v3.2 to 71.3 on BrowseComp-ZH and 44.4 on HLE, surpassing several proprietary foundation models, and establishes leading performance for Tongyi-DR-30B-A3B among information-seeking agents of comparable scale.

Our core contributions are as follows:

\begin{itemize}
    \item We introduce the first probabilistic framework for long-horizon web agents that characterizes context management through two complementary dimensions, search efficiency $\eta$ and terminal precision $\rho$, providing a unified lens for understanding the behavior of different strategies.
    
    \item We propose AgentSwing, a state-aware adaptive context management framework that dynamically switches among candidate strategies according to the quality of the current trajectory and continuations, thereby balancing search efficiency and terminal reliability and improving overall long-horizon agent performance.
    
    \item Extensive experiments across multiple long-horizon benchmarks and model backbones demonstrate the effectiveness and generalization of AgentSwing, and provide a fine-grained analysis of how different context management strategies behave and why adaptive routing works.
\end{itemize}

\section{A Complementary Probabilistic View of Long-Horizon Web Agents}

We begin with a probabilistic characterization of long-horizon web agents under resource-constrained execution. In deep information-seeking, end-to-end success cannot be understood solely by final answer accuracy. Before producing a correct answer, the agent must first navigate a long interaction trajectory, accumulate sufficient evidence, and reach a stopping point before exhausting its available resources, such as context budget and maximum interaction turns. Accordingly, failures arise from two distinct sources: the agent may fail to reach a stopping point within the allowed resources, or it may terminate but produce an incorrect answer.

\subsection{Two Perspectives on Success: Search Efficiency and Terminal Precision}

We assume tasks $\tau$ are independently sampled from an underlying task distribution $\mathcal{T}$. 
For a task $\tau$, consider an agent executed under a test-time strategy $\pi$, where $\pi$ specifies the execution protocol, including context management, stopping rules, and resource constraints. Let $S^\pi$ denote the event that the agent reaches a stopping point and emits a final answer under strategy $\pi$, and let $C^\pi$ denote the event that this answer is correct. 
We define two task-level quantities:

\begin{equation}
\eta_\tau^\pi := P(S^\pi \mid \tau), \quad 
\rho_\tau^\pi := P(C^\pi \mid S^\pi, \tau).
\end{equation}

Here, $\eta_\tau^\pi$ is the agent's \emph{search efficiency}, i.e., the probability of reaching a stopping point before the protocol terminates, and $\rho_\tau^\pi$ is its \emph{terminal precision}, i.e., the probability that the answer is correct conditioned on reaching such a stopping point.

Task-level success then follows from the chain rule:

\begin{equation}
P(\mathrm{Success}^\pi \mid \tau)
= P(S^\pi \cap C^\pi \mid \tau)
= \eta_\tau^\pi \rho_\tau^\pi.
\label{eq:task_success_factorization}
\end{equation}

Thus, success requires both reaching a terminal state and answering correctly once there. At the population level, we define

\begin{equation}
\eta^\pi := P(S^\pi) = \mathbb{E}_{\tau \sim \mathcal{T}}[\eta_\tau^\pi],
\label{eq:eta_population}
\end{equation}

\begin{equation}
\rho^\pi := P(C^\pi \mid S^\pi)
= \frac{P(C^\pi \cap S^\pi)}{P(S^\pi)}
= \frac{\mathbb{E}_{\tau \sim \mathcal{T}}[\eta_\tau^\pi \rho_\tau^\pi]}
{\mathbb{E}_{\tau \sim \mathcal{T}}[\eta_\tau^\pi]}.
\label{eq:rho_population}
\end{equation}

Accordingly, the population-level success probability can be written as

\begin{equation}
\mathrm{Pass@1}^\pi = P(\mathrm{Success}^\pi) =P(S^\pi \cap C^\pi)= \eta^\pi \rho^\pi.
\label{eq:population_factorization}
\end{equation}

This decomposition shows that commonly used end-to-end metrics such as Pass@1 or accuracy should not be treated as monolithic indicators in long-horizon settings. Instead, they jointly reflect search efficiency and terminal precision.

In practice, suppose a benchmark contains $M$ tasks. For a fixed strategy $\pi$, let $N^\pi_{\mathrm{finish}}$ denote the number of tasks on which the agent reaches a stopping point and emits a final answer, and let $N^\pi_{\mathrm{correct}}$ denote the number of tasks on which the final answer is correct. Following \citet{team2026kimik2.5, zeng2026glm}, tasks that exhaust the allowed resources before producing a final answer are directly counted as failed. We estimate

\begin{equation}
\eta^\pi \approx \frac{N^\pi_{\mathrm{finish}}}{M},
\qquad
\rho^\pi \approx \frac{N^\pi_{\mathrm{correct}}}{N^\pi_{\mathrm{finish}}},
\label{eq:empirical_estimators}
\end{equation}

with the corresponding empirical end-to-end success rate

\begin{equation}
\mathrm{Pass@1}^\pi
=
\eta^\pi \rho^\pi
\approx
\frac{N^\pi_{\mathrm{correct}}}{M}.
\label{eq:empirical_pass1}
\end{equation}

Since different strategies may finish on different task subsets, we additionally report \emph{aligned terminal precision} for cross-strategy comparison. Let $N_{\mathrm{aligned\mbox{-}finish}}$ be the number of tasks that finish under all compared strategies or settings, and let $N^\pi_{\mathrm{aligned\mbox{-}correct}}$ be the number of these tasks answered correctly by strategy $\pi$. We compute

\begin{equation}
\rho^\pi_{\mathrm{align}}
\approx
\frac{N^\pi_{\mathrm{aligned\mbox{-}correct}}}{N_{\mathrm{aligned\mbox{-}finish}}}.
\label{eq:aligned_precision}
\end{equation}

By reporting terminal precision on the shared finished subset, this metric enables a fairer comparison across strategies or settings.

\subsection{Discard-All vs. Baseline}
\label{sec:trade-off-2.2}
We use \emph{Discard-All} as a concrete case study to instantiate the framework above and explain why context management can outperform the standard \emph{w/o context management} baseline.

Let $\pi=\mathrm{std}$ denote the baseline without context management. Under this protocol, the agent continuously appends its interaction history and follows a single uninterrupted search trajectory. It therefore either reaches a stopping point and produces a final answer, or exhausts the maximum context length and is counted as failed. In contrast, \emph{Discard-All} ($\pi=\mathrm{DA}$) introduces a context-management trigger. Once the accumulated context exceeds a predefined threshold, the agent discards the full trajectory history and continues from the original user prompt only. As a result, the same task execution under \emph{Discard-All} may contain multiple reset-based attempts. If the maximum turn budget is exhausted before a final answer is produced, the task is counted as failed.

\begin{wrapfigure}{r}{8.6cm}
    \vspace{-2em}
    \centering
    \includegraphics[width=0.95\linewidth]{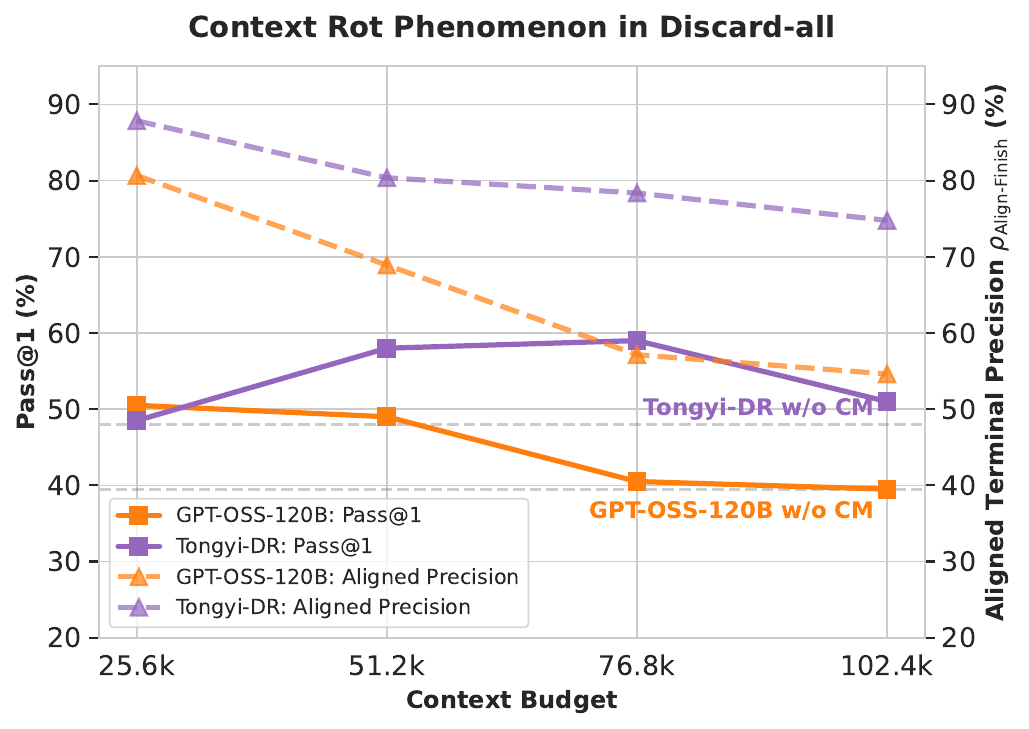}
    \caption{Performance on BrowseComp under \textit{Discard-All} with different context budgets.}
    \vspace{-1.2em} 
    \label{fig:context_rot}
\end{wrapfigure}

We next study how the trigger threshold affects the performance of \emph{Discard-All}, while also using it to understand the difference between \emph{Discard-All} and the baseline. We vary the trigger ratio while fixing the maximum interaction turns to 400, so that the primary changing factor is the effective context budget per attempt. Figure~\ref{fig:context_rot} shows that, for both Tongyi-DR-30B-A3B and GPT-OSS-120B, aligned terminal precision consistently decreases as the context budget increases. This indicates that larger working contexts lead to more severe context rot at termination~\citep{hsieh2024ruler,modarressi2025nolima,hong2025context,fang2026agentlongbench}. Since the baseline corresponds to the largest context regime, it is also the least favorable for terminal precision. At the same time, an appropriate context budget allows \emph{Discard-All} to outperform the baseline in overall performance.

This phenomenon can be further interpreted through our efficiency-precision framework. In Figure~\ref{fig:cm_compare_3panel}b, the standard baseline typically has the lowest terminal precision, consistent with the trend in Figure~\ref{fig:context_rot}, but also the highest search efficiency. In other words, it reaches stopping points on more tasks, yet the resulting terminal states are less reliable. By contrast, \emph{Discard-All} usually has lower search efficiency $\eta$, because each reset-based attempt operates under a smaller effective context budget and is less likely to finish on its own. However, this efficiency loss can be alleviated by increasing the number of reset opportunities $N$. For a task $\tau$, let $S_i$ denote the event that the agent reaches a stopping point during the $i$-th reset-based attempt, and suppose at most $N$ such attempts are allowed. Then
\begin{equation}
\eta_\tau^{\mathrm{DA}}
=
P\Bigl(\bigcup_{i=1}^{N} S_i \,\Big|\, \tau\Bigr),
\label{eq:discard_eta_general}
\end{equation}
which, under a conditional independence approximation across reset-based segments, becomes
\begin{equation}
\eta_\tau^{\mathrm{DA}}
=
1 - \prod_{i=1}^{N}\bigl(1-\eta_{\tau,i}^{\mathrm{DA}}\bigr)
\approx
1 - \bigl(1-\eta_{\tau,\mathrm{single}}^{\mathrm{DA}}\bigr)^N.
\label{eq:discard_eta_indep}
\end{equation}
Although each individual attempt is less likely to finish than the baseline, increasing $N$ provides more chances to reach a stopping point. Combined with the higher precision of smaller contexts, this allows \emph{Discard-All} to outperform the baseline.

\subsection{Static Context Management Strategies in the Efficiency-Precision Plane}

\label{sec2:trade-off}

\begin{figure*}[ht]
    \centering
    \includegraphics[width=\linewidth]{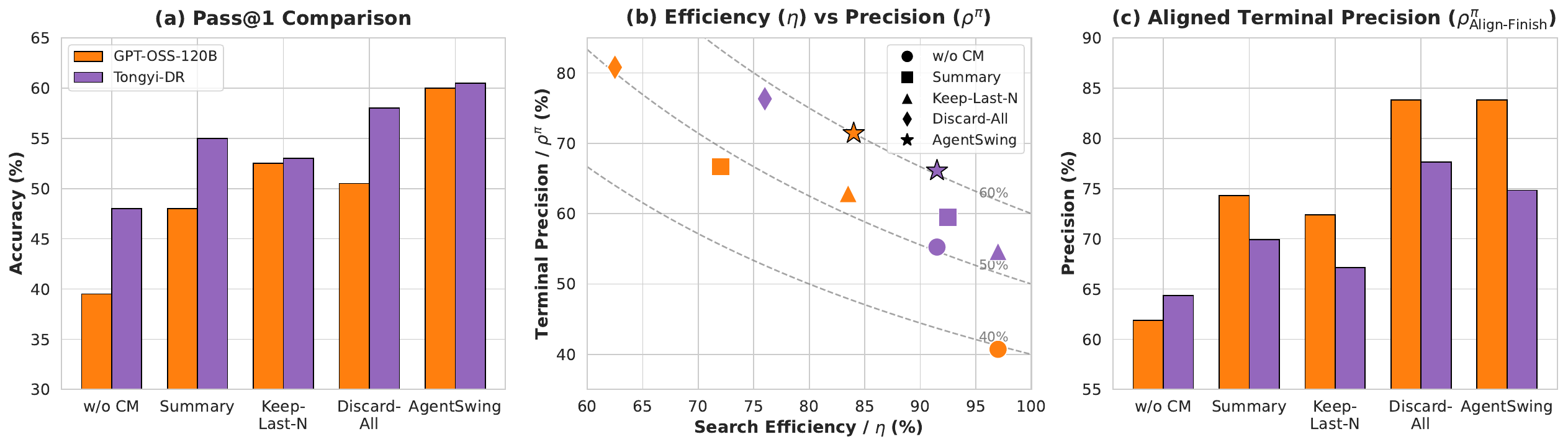}
    \caption{Comparison of context management strategies through the efficiency-precision lens. (a) Pass@1 on BrowseComp. (b) Search efficiency $\eta$ vs. terminal precision $\rho$. (c) Aligned terminal precision, where \emph{Align-Finish} refers to the common finished cases shared by different strategies within the same model.}
    \vspace{-3mm}
    \label{fig:cm_compare_3panel}
\end{figure*}

The same perspective extends naturally beyond \emph{Discard-All} to other context management strategies. Figure~\ref{fig:cm_compare_3panel} compares \emph{Summary}, \emph{Discard-All}, \emph{Keep-Last-N}, and \emph{AgentSwing} under maximum interaction budget of 400 turns. As shown in Figure~\ref{fig:cm_compare_3panel}a, all context management strategies outperform the baseline in Pass@1, but through different efficiency-precision trade-offs.

Figure~\ref{fig:cm_compare_3panel}b shows that static strategies occupy different operating points in the efficiency-precision plane, forming an empirical trade-off frontier. As discussed above, the standard baseline is high-efficiency but low-precision, whereas \emph{Discard-All} lies near the opposite end. \emph{Summary} and \emph{Keep-Last-N} fall between these extremes, improving search efficiency over \emph{Discard-All} but not matching its terminal precision; see also Figure~\ref{fig:cm_compare_3panel}c. In contrast, \emph{AgentSwing} moves to a more favorable region of the plane by adaptively routing multiple strategies, leading to the strongest overall performance.

\begin{figure*}[ht]
    \centering
    \includegraphics[width=\textwidth]{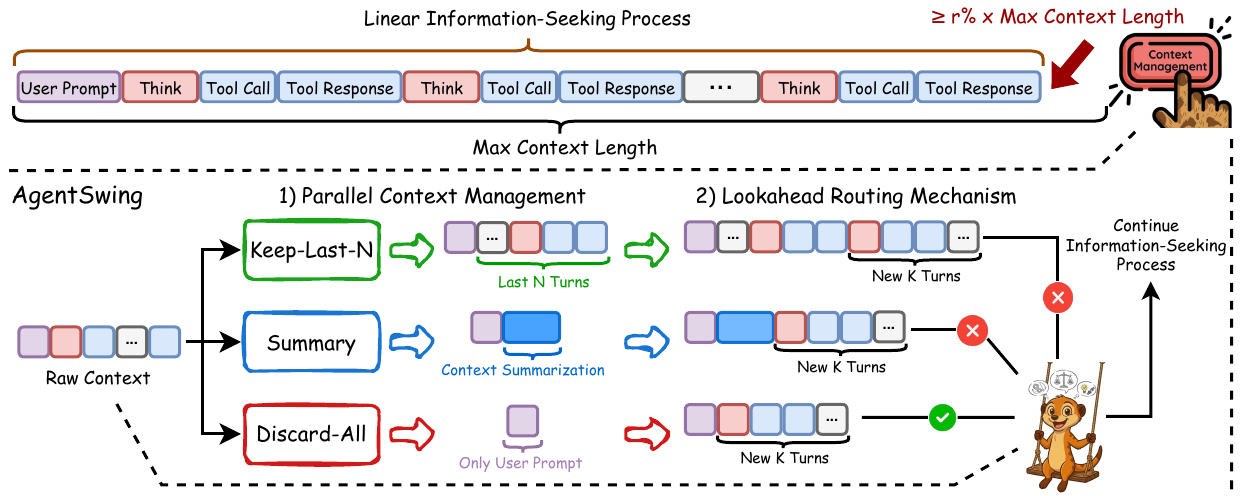}
    \caption{Overview of AgentSwing. AgentSwing triggers context management once the accumulated context exceeds a predefined threshold, executes multiple candidate strategies in parallel, extends each branch for $K$ new turns, and dynamically routes to the most promising continuation.} 
    \vspace{-3mm}
    \label{fig:agentswing_method}
\end{figure*}

\section{AgentSwing}

AgentSwing consists of two components: (1) \emph{Parallel Context Management} and (2) \emph{Lookahead Routing}, as illustrated in Figure~\ref{fig:agentswing_method}. We consider the standard deep information-seeking setting, where an agent starts from a user prompt $q$ and interacts with the environment through repeated \texttt{(<thinking>, <tool call>, <tool response>)} turns. When the current context length exceeds a predefined fraction $r$ of the model's maximum context length, the framework activates context management over the accumulated trajectory.

\noindent{\textbf{(1) Parallel Context Management.}}\label{sec:parallel_cm_methods}
At each trigger point, AgentSwing applies multiple candidate context management strategies to the same raw context in parallel, producing a set of alternative managed contexts. In this work, we consider three representative strategies:

\begin{itemize}
   \vspace{-2mm}
    \item \textbf{Keep-Last-N:} Preserves only the latest $N$ interaction turns, i.e., the last $N$ \texttt{(<thinking>, <tool call>, <tool response>)} tuples, and discards earlier history~\citep{liu2025deepseek,zeng2026glm}.
   \vspace{-2mm}
    \item \textbf{Summary:} Compresses the accumulated trajectory into a summarized text and keeps the context in the form of the original user prompt together with the summary, i.e., $(q,\mathrm{Sum})$~\citep{liu2025deepseek,context-engineering}.
    \vspace{-2mm}
    \item \textbf{Discard-All:} Discards the entire accumulated interaction history and keeps only the original user prompt $q$~\citep{liu2025deepseek,team2026kimik2.5,zeng2026glm}.
   \vspace{-1mm}
\end{itemize}

Applying these strategies in parallel can further yield multiple candidate continuations from the same trajectory state, each corresponding to a different way of managing the accumulated context.

\noindent{\textbf{(2) Lookahead Routing Mechanism.}}
After parallel context management, AgentSwing does not immediately select a branch. 
Instead, it performs short-horizon lookahead for each managed context. Concretely, each branch continues interacting with the environment for $K$ additional turns.
After the lookahead phase, AgentSwing presents the candidate continuations together with the original raw context to the agent model, which then selects the most reasonable branch for subsequent exploration. The remaining branches are discarded, and the selected continuation becomes the new main trajectory.
This design allows branch selection to depend not only on the managed context itself, but also on its short-term downstream behavior under real environment feedback. In this way, AgentSwing differs from static strategies, which repeatedly apply a single fixed strategy throughout the entire search process.

\section{Experiments}

\subsection{Setup}\label{sec:setup}

\noindent{\textbf{Benchmarks.}} 
We evaluate AgentSwing on three challenging deep information-seeking benchmarks: BrowseComp~\citep{bc_en}, BrowseComp-ZH~\citep{bc_zh}, and Humanity's Last Exam (HLE)~\citep{hle}. 
These benchmarks jointly assess deep search and reasoning ability. For efficient evaluation, we use sampled subsets for the larger benchmarks: 200 randomly selected tasks from BrowseComp and 500 text-only tasks from HLE, following prior work~\citep{Li2025webthinker,nguyen2025sfr}. 
For BrowseComp-ZH, we use the full set of 289 tasks.

\noindent{\textbf{Tools.}} 
We adopt the standard tool configuration used by deep information-seeking agents~\citep{wu2025webdancerautonomousinformationseeking,li2025websailorv2bridgingchasmproprietary}, with Search and Visit as the core tools. For HLE, following \citet{chen2026iterresearch}, we further include Google Scholar and a Python Interpreter. Details are as follows:

\begin{itemize}
    \vspace{-0.5em}
    \item \textbf{Search:} Performs batched Google queries and returns the top-10 ranked results for each query.
    \item \textbf{Visit:} Fetches webpages from  URLs and extracts information relevant to the specified goal.
    \item \textbf{Google Scholar:} Returns top-10 academic search results with snippets, citations, and scholarly metadata.
    \item \textbf{Python Interpreter:} Executes arbitrary Python code in a secure sandbox for computational tasks and data analysis. We use Code Sandbox\footnote{\href{https://github.com/bytedance/SandboxFusion}{https://github.com/bytedance/SandboxFusion}} to ensure secure and isolated execution.
\end{itemize}

\noindent{\textbf{Agent Models.}} 
We use three open-source models with diverse parameter scales and strong tool-use capability for deep information-seeking tasks: GPT-OSS-120B~\citep{gptoss}, DeepSeek-v3.2~\citep{liu2025deepseek}, and Tongyi-DeepResearch-30B-A3B (Tongyi-DR-30B-A3B)~\citep{tongyidr}. 
All models are invoked under their official function-calling protocol. 
Unless otherwise specified, we use the same agent model for both stages in AgentSwing.

\begin{table*}[ht]
\small
\centering
\caption{Overall performance on long-horizon agentic benchmarks. Scores marked with $\ddag$ represent full-benchmark results, whereas unmarked scores correspond to our benchmark settings. }
\label{tab:main_result}
\resizebox{1.0\textwidth}{!}{
\setlength{\tabcolsep}{6pt}
\renewcommand{\arraystretch}{1.2}
\begin{tabular}{l|l|c|c|c}
\toprule
\textbf{Model / Framework} & \textbf{Context Management} & \textbf{BrowseComp} & \textbf{BrowseComp-ZH} & \textbf{HLE} \\
\midrule
\multicolumn{5}{c}{\cellcolor{blue!20}\textit{\textbf{Foundation Models with Tools}}} \\
\midrule
Claude-4.5-Opus~\citep{claude4.5opus} & w/o CM & 37.0$^\ddag$ & 62.4 & 43.4$^\ddag$ \\
Gemini-3.0-Pro~\citep{gemini3.0} & w/o CM & 37.8$^\ddag$ & 66.8 & 45.8$^\ddag$ \\
GPT-5.1 High~\citep{gpt5.1} & w/o CM & 50.8$^\ddag$ & -- & 42.7$^\ddag$ \\
OpenAI-o3~\citep{o3} & w/o CM & 49.7$^\ddag$ & 58.1 & -- \\
\midrule
\multicolumn{5}{c}{\cellcolor{blue!20}\textit{\textbf{Deep Information-Seeking Agents}}} \\
\midrule
OpenAI DeepResearch~\citep{openaidr} & - & 51.5$^\ddag$ & 42.9 & 26.6$^\ddag$ \\
ASearcher-Web-32B~\citep{asearcher} & - & 5.2$^\ddag$ & 15.6 & 12.5$^\ddag$ \\
DeepMiner-32B-RL~\citep{deepminer} & - & 33.5$^\ddag$ & 40.1 & -- \\
IterResearch-30B-A3B~\citep{chen2026iterresearch} & - & 37.3$^\ddag$ & 45.2 & 28.8$^\ddag$ \\
AgentFold-30B-A3B~\citep{ye2026agentfold} & - & 36.2$^\ddag$ & 47.3 & -- \\
AgentFounder-30B-A3B~\citep{su2026agentfounder} & - & 39.9$^\ddag$ & 43.3 & 31.5$^\ddag$ \\
MiroThinker-v1.5-30B-A3B~\citep{mirothinkerv1.5} & - & 56.1$^\ddag$ & 66.8 & 31.0$^\ddag$ \\
\midrule
\multicolumn{5}{c}{\cellcolor{blue!30}\textit{\textbf{Open-Source Models with Context Management}}} \\
\midrule
\multirow{5}{*}{GPT-OSS-120B} & Baseline (w/o CM) & 39.5 & 28.4 & 33.2 \\
& Discard-All & 50.5 & 31.5 & 34.2 \\
& Keep-Last-N & 52.5 & 33.6 & 34.1 \\
& Summary & 48.0 & 30.8 & 34.4 \\
& \cellcolor{blue!10}\textbf{AgentSwing (Ours)} & \cellcolor{blue!10}\textbf{60.0} & \cellcolor{blue!10}\textbf{38.0} & \cellcolor{blue!10}\textbf{35.1} \\
\midrule
\multirow{5}{*}{DeepSeek-v3.2} & Baseline (w/o CM) & 51.4$^\ddag$ / 43.5 & 65.0$^\ddag$ / 61.6 & 40.8$^\ddag$ / 40.2 \\
& Discard-All & 58.0 & 70.2 & 42.0 \\
& Keep-Last-N & 52.0 & 69.9 & 39.6 \\
& Summary & 48.5 & 69.2 & 43.5 \\
& \cellcolor{blue!10}\textbf{AgentSwing (Ours)} & \cellcolor{blue!10}\textbf{62.5} & \cellcolor{blue!10}\textbf{71.3} & \cellcolor{blue!10}\textbf{44.4} \\
\midrule
\multirow{5}{*}{Tongyi-DR-30B-A3B} & Baseline (w/o CM) & 43.4$^\ddag$ / 48.0 & 46.7$^\ddag$ / 47.1 & 32.9$^\ddag$ / 31.7 \\
& Discard-All & 58.0 & 53.9 & 32.7 \\
& Keep-Last-N & 53.0 & 50.1 & 32.2 \\
& Summary & 55.0 & 49.1 & 32.0 \\
& \cellcolor{blue!10}\textbf{AgentSwing (Ours)} & \cellcolor{blue!10}\textbf{60.5} & \cellcolor{blue!10}\textbf{56.7} & \cellcolor{blue!10}\textbf{33.1} \\
\bottomrule
\end{tabular}
}
\end{table*}

\noindent{\textbf{Baselines.}} 
In addition to the standard baseline without context management (\textit{w/o CM}), we compare AgentSwing with several representative static context management strategies introduced in Section~\ref{sec:parallel_cm_methods}, including \textit{Discard-All}, \textit{Keep-Last-N} ($N=5$), and \textit{Summary}. 
For \textit{Summary}, the summarization step is always performed by GPT-OSS-120B. 

\noindent{\textbf{Evaluation Metrics and Hyper-parameters.}}\label{app:parameters} 
All evaluations are conducted under the LLM-as-a-Judge protocol~\citep{llmasajudge}, using the official evaluation prompts and judging models released by each benchmark. 
For all agent models, we set the maximum context length to 128k tokens. Unless otherwise specified, we set the maximum interaction budget to 400 turns for all context management strategies. To ensure fair comparison and reproducibility, model-specific hyper-parameters follow the officially recommended or empirically optimal settings of each agent backbone. 
For all experiments involving context management, we set the context budget as a fixed ratio $r$ of the 128k maximum context length. 
Specifically, we use $r=0.2$ for GPT-OSS-120B and $r=0.4$ for both Tongyi-DR-30B-A3B and DeepSeek-v3.2. 
The rationale behind these settings is discussed in Section~\ref{sec:trade-off-2.2}.

\subsection{Overall Performance}

Table~\ref{tab:main_result} shows that AgentSwing consistently achieves advanced performance across all benchmarks and agent backbones, outperforming both the standard baseline and representative context management strategies. Notably, AgentSwing pushes DeepSeek-v3.2 to 71.3 on BrowseComp-ZH and 44.4 on HLE, surpassing several proprietary foundation models. It also establishes leading performance for Tongyi-DR-30B-A3B among deep information-seeking agents of comparable scale. These results show that adaptive context management is a strong and general test-time scaling mechanism for long-horizon web agents.

\subsection{Analysis and Ablation}
We next provide a fine-grained analysis of AgentSwing. We examine how different context management strategies scale with interaction budget, compare their behavior on aligned harder cases, ablate the lookahead routing mechanism, and present case studies. Further analyses of strategy combinations and strategy transitions are deferred to Appendices~\ref{app:cm_combi} and \ref{app:trans}.

\begin{figure*}[ht]
    \centering
    \includegraphics[width=\linewidth]{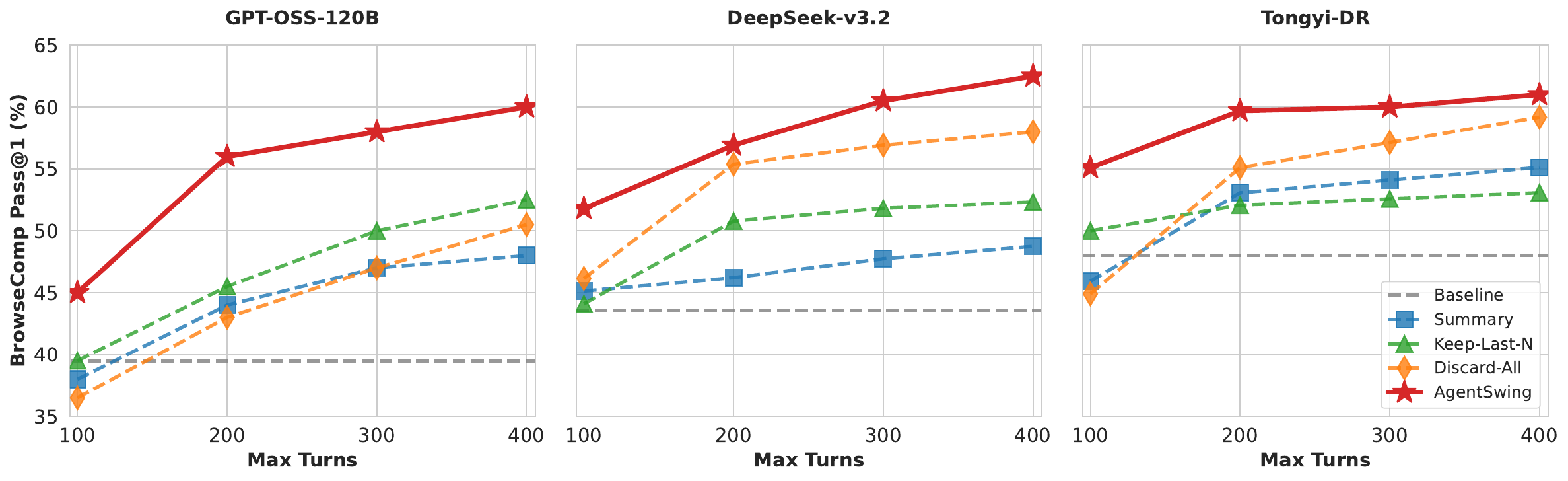}
    \caption{Performance of different context management strategies on BrowseComp over maximum interaction turns.} 
    \vspace{-3mm}
    \label{fig:pass1_over_turns}
\end{figure*}

\noindent\textbf{Analysis of Context Management Strategies.}
Figure~\ref{fig:pass1_over_turns} shows how different context management strategies scale with the maximum interaction budget on BrowseComp. Under small turn budgets, context management provides only limited gains over the baseline, and some static strategies may even 

\begin{table*}[ht]
\small
\centering
\caption{Performance comparison of different context management methods on aligned cases that trigger context management under all evaluated strategies ($\rho_{\text{Align-CM}}$).}
\label{tab:aligned_context_management}
\resizebox{1.0\textwidth}{!}{
\setlength{\tabcolsep}{8pt}
\renewcommand{\arraystretch}{1.2}
\begin{tabular}{l|l|c|c|c|c|c|c|c}
\toprule
\textbf{Model} & \textbf{Strategy} & $\mathbf{N_{\text{align}}}$ & $\mathbf{N_{\text{finish}}}$ & $\mathbf{N_{\text{correct}}}$ & \textbf{$\eta$ (\%) }& \textbf{$\rho$ (\%)} & \textbf{Pass@1 (\%)} & $\mathbf{\overline{N_{\text{turn}}}}$ \\
\midrule
\multirow{4}{*}{GPT-OSS-120B} 
& Discard-All & \multirow{4}{*}{122} & 51 & 35 & 41.8 & \cellcolor{blue!10}\textbf{68.6} & 28.7 & 297.2 \\
& Summary & & 68 & 35 & 55.7 & 51.5 & 28.7 & 248.0 \\
& Keep-Last-N & & 91 & 43 & \cellcolor{blue!10}\textbf{74.6} & 47.3 & 35.2 & 205.4 \\
& AgentSwing & & 90 & 51 & 73.8 & 56.7 & \cellcolor{blue!10}\textbf{41.8} & \cellcolor{blue!10}\textbf{190.3} \\
\midrule
\multirow{4}{*}{DeepSeek-v3.2} 
& Discard-All & \multirow{4}{*}{73} & 40 & 24 & 54.8 & \cellcolor{blue!10}\textbf{60.0} & 32.9 & 268.3 \\
& Summary & & 72 & 22 & \cellcolor{blue!10}\textbf{98.6} & 30.6 & 30.1 & \cellcolor{blue!10}\textbf{132.2} \\
& Keep-Last-N & & 53 & 23 & 72.6 & 43.4 & 31.5 & 183.5 \\
& AgentSwing & & 68 & 26 & 93.2 & 38.2 & \cellcolor{blue!10}\textbf{35.6} & 151.9 \\
\midrule
\multirow{4}{*}{Tongyi-DR-30B-A3B} 
& Discard-All & \multirow{4}{*}{45} & 11 & 9 & 24.4 & \cellcolor{blue!10}\textbf{81.8} & 20.0 & 340.8 \\
& Summary & & 35 & 9 & 77.8 & 25.7 & 20.0 & 215.7 \\
& Keep-Last-N & & 42 & 9 & \cellcolor{blue!10}\textbf{93.3} & 21.4 & 20.0 & \cellcolor{blue!10}\textbf{153.0} \\
& AgentSwing & & 34 & 14 & 75.6 & 41.2 & \cellcolor{blue!10}\textbf{31.1} & 203.6 \\
\bottomrule
\end{tabular}
}
\vspace{-2mm}
\end{table*}

underperform it, since the baseline benefits from its large single-attempt context and therefore retains relatively strong search efficiency. Once the budget becomes sufficiently large, all context management strategies consistently surpass the baseline, indicating that the precision advantage of managed contexts becomes dominant as more interaction turns are allowed. This trend matches the analysis in Section~\ref{sec2:trade-off}. AgentSwing stands out by outperforming the baseline even under limited budgets and maintaining a consistent advantage over static strategies across the full scaling curve.

To further isolate strategy behavior on harder cases, Table~\ref{tab:aligned_context_management} reports results on the subset of tasks where context management is triggered under all compared strategies within the same model. We can observe that \textit{Keep-Last-N} and \textit{Summary} usually achieve stronger search efficiency $\eta$, while \textit{Discard-All} achieves the strongest terminal precision $\rho$. AgentSwing combines the strengths of both regimes, with efficiency close to the former and precision close to the latter, leading to the highest overall Pass@1 across all three models on this aligned subset. Moreover, AgentSwing also achieves average turn counts close to the more efficiency-oriented strategies, while being substantially more efficient than \textit{Discard-All}. This shows that its gains do not come from simply paying a larger interaction cost, but from adaptively selecting the most suitable context management decision according to the current trajectory state.

\noindent\textbf{Ablation of the Lookahead Routing Mechanism.} To validate the effectiveness of the routing mechanism, \begin{wraptable}{r}{8cm}
\small
\centering
\vspace{-2mm}
\caption{Ablation on lookahead strategy.}
\label{tab:lookahead_ablation}
\resizebox{\linewidth}{!}{
\setlength{\tabcolsep}{6pt}
\renewcommand{\arraystretch}{1.2}
\begin{tabular}{l|c|c}
\toprule
\textbf{Routing Mechanism} & \textbf{GPT-OSS-120B} & \textbf{Tongyi-DR-30B-A3B} \\
\midrule
random & 51.0 & 56.5 \\
w/o Lookahead & 50.0 & 57.0 \\
Lookahead ($k=1$) & 52.5 & 58.0 \\
Lookahead ($k=3$) & \cellcolor{blue!10}\textbf{60.0} & \cellcolor{blue!10}\textbf{60.5} \\
Lookahead ($k=5$) & 55.0 & 59.0 \\
\bottomrule
\end{tabular}
}
\vspace{-2mm}
\end{wraptable}we report ablations in Table~\ref{tab:lookahead_ablation}. We compare AgentSwing with two variants: \textit{random}, which selects a context management branch uniformly at random after triggering, and \textit{w/o Lookahead}, which performs parallel context management but removes rollout and therefore selects solely based on the managed contexts themselves. Both variants consistently underperform AgentSwing, showing that the gains do not come merely from maintaining multiple candidate strategies, but from using short-horizon lookahead to evaluate their downstream consequences before routing.

We further vary the lookahead depth $k$, i.e., the number of newly generated turns per branch before routing. 
The results show that moderate lookahead is most effective. In particular, $k=3$ generally provides the strongest performance across models. Compared with $k=1$, it exposes richer future trajectory information, while larger lookahead such as $k=5$ does not always improve performance further, since it may risk exceeding maximum length constraints of agent models.

\begin{figure*}[ht]
    \centering
    \includegraphics[scale=0.38]{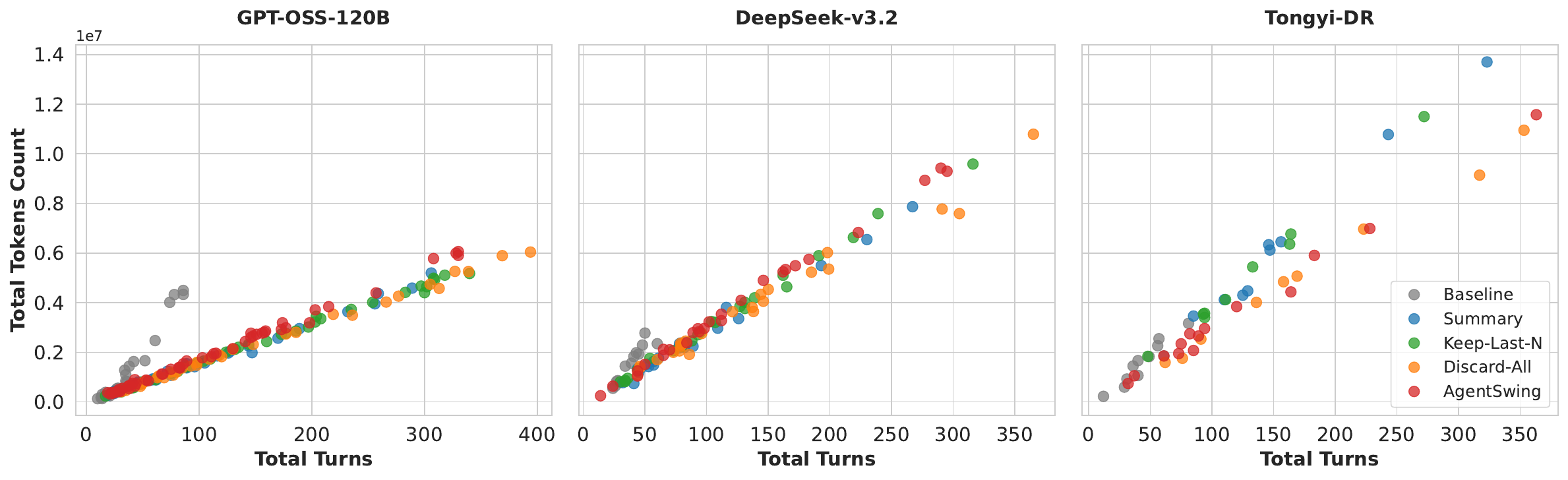}
    \caption{Token efficiency on the aligned cases used in Table~\ref{tab:aligned_context_management}. Each point corresponds to one finished task and is plotted by its total interaction turns and cumulative token count at termination.}
    \vspace{-3mm}
    \label{fig:token_count}
\end{figure*}

\noindent\textbf{Comparison of Token Efficiency.} 
Figure~\ref{fig:token_count} compares token efficiency on the aligned cases used in Table~\ref{tab:aligned_context_management}. Each point denotes one finished task, plotted by its total interaction turns and cumulative token count at termination. Although \textit{AgentSwing} introduces additional token usage due to lookahead routing, the overhead remains modest in practice. One reason is that efficiency-oriented strategies such as \textit{Keep-Last-N} often incur higher cumulative token usage at similar turn counts, since they retain more trajectory history in the context. By contrast, \textit{Discard-All} tends to accumulate fewer tokens, but usually requires more turns to finish. Taken together, these results show that AgentSwing does not achieve its gains by paying a substantially larger overall cost.

\subsection{Case Study}
\label{sec:case}
\begin{figure*}[ht]
    \centering
    \includegraphics[width=\linewidth]{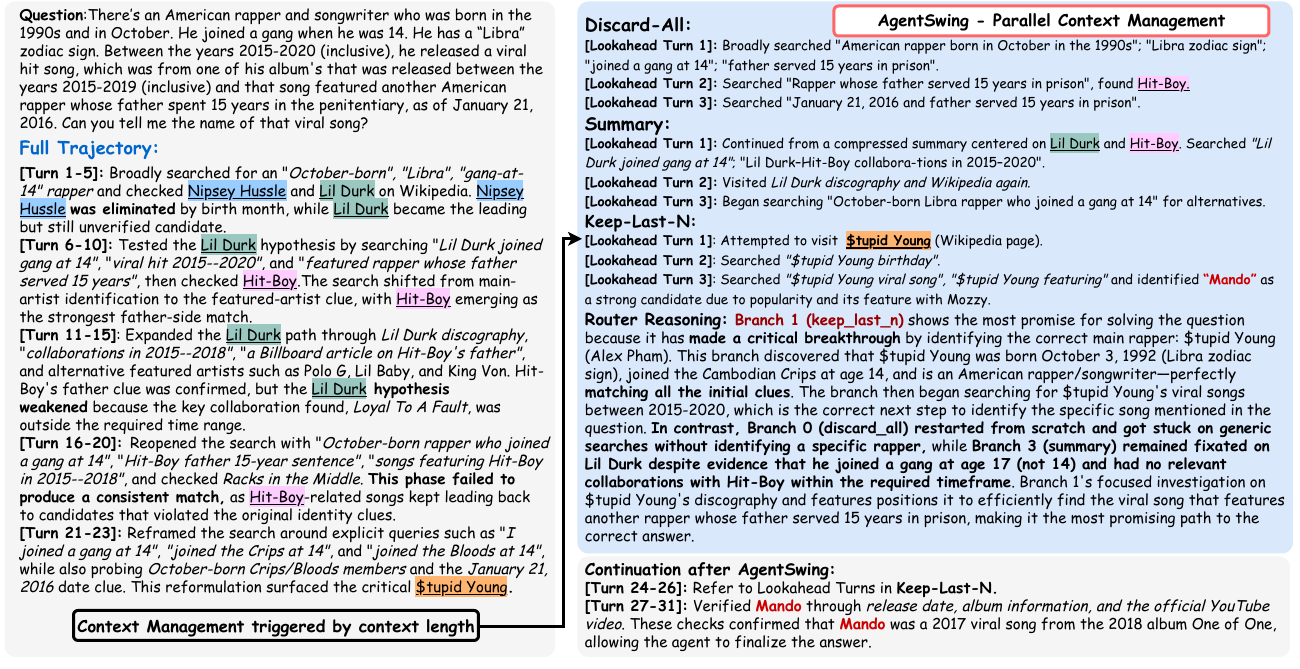}
    \caption{Case Study.} 
    \vspace{-3mm}
    \label{fig:case_study}
\end{figure*}

Figure~\ref{fig:case_study} shows a case from DeepSeek-v3.2. When context management is triggered at Turn 23, the current history contains both substantial distractions arising from incorrect hypotheses (``Nipsey Hussle'', ``Lil Durk'', and ``Hit-Boy'') and a newly surfaced local clue (``\$tupid Young''). This mixed state makes static context management brittle.

The three candidate branches produce clearly different continuations. \textit{Discard-All} resets the search and falls back to broad exploration. \textit{Summary} preserves the dominant but incorrect ``Lil Durk'' hypothesis. In contrast, \textit{Keep-Last-N} preserves the recent clue chain around ``\$tupid Young'', enabling the agent to verify the rapper's identity and identify ``Mando'' within lookahead turns. Based on these continuations, the router correctly selects \textit{Keep-Last-N}, after which the agent verifies the remaining constraints and reaches the final answer shortly afterward. This example illustrates the central advantage of AgentSwing. It treats context management as a state-dependent routing problem over future continuations rather than as a fixed compression heuristic. Appendix~\ref{app:case_study} provides a detailed turn-by-turn summary of this example, together with a complementary GPT-OSS-120B case in which \textit{Discard-All} is selected.

\section{Related Work}

\noindent\textbf{Long-horizon web agents.}
LLM-based web agents have rapidly evolved from single-turn assistants into autonomous systems capable of web browsing, tool use, and long-horizon information seeking~\citep{webwalker,wu2025webdancerautonomousinformationseeking,li2025websailornavigatingsuperhumanreasoning,fang2025towards,liu2025webexplorer}. 
Recent efforts from both academia and industry have demonstrated strong potential on deep information-seeking tasks, while also highlighting the importance of test-time scaling and long-horizon interaction design~\citep{chai2025scimaster,huang2025environment,li2025parallelmuse,zeng2026pushing}.
However, most existing agents still rely on ReAct-style trajectories~\citep{yao2023react}, making them increasingly vulnerable to context saturation, drift, and error accumulation as the search horizon grows~\citep{fang2026agentlongbench}.

\noindent\textbf{Context management for LLM agents.}
Context management, or context engineering, aims to provide LLM-based agents with a more effective working context~\citep{context-engineering,qiao2025webresearcher}. 
Within long-horizon agents, prior methods mainly rely on static intra-task context curation, including reset-based policies such as \textit{Discard-All}, recent-turn retention such as \textit{Keep-Last-N}~\citep{liu2025deepseek,team2026kimik2.5,zeng2026glm}, and context compaction strategies closely related to \textit{Summary}~\citep{yu2025memagent,ye2026agentfold,context-engineering,liu2025deepseek}. 
These methods improve context efficiency, but once a strategy is selected, the same operation is repeatedly applied throughout the entire trajectory. In contrast, AgentSwing treats context management as a state-dependent routing problem and dynamically selects among heterogeneous strategies.

\section{Conclusion}

In this work, we introduce the first probabilistic framework that decomposes the end-to-end success of deep information-seeking agents into two complementary dimensions, search efficiency and terminal precision, providing a unified view of how context management strategies affect long-horizon performance. Building on this perspective, we propose AgentSwing, an adaptive framework that moves beyond a single static context management strategy by expanding multiple parallel context management branches and dynamically selecting among them through a lookahead routing mechanism.
Experiments across multiple benchmarks and backbones demonstrate that AgentSwing is both effective and generalizable, consistently improving long-horizon agent performance over static context management baselines.

\section{Limitations and Future Work}

Our work focuses on test-time context management as an external control mechanism for long-horizon agents. 
The proposed perspective helps clarify the efficiency-precision trade-off and leads to strong empirical gains. 
A more fundamental direction is to translate these principles into model-level competence, for example, by training agents that are intrinsically more efficient under smaller context budgets or more reliable under long-horizon noisy trajectories. In addition, the current routing mechanism is still performed by the agent model itself. 
Although this design is simple and effective, it may not be optimal. 
A stronger dedicated router, verifier, or trajectory evaluator with better foresight may further improve branch selection quality and therefore unlock additional gains for adaptive context management.

\clearpage

\clearpage
\bibliography{custom}
\bibliographystyle{colm2024_conference}

\clearpage
\appendix

\section{Gains from Parallel Context Management Combinations}
\label{app:cm_combi}
\begin{wrapfigure}{r}{8cm}
    \centering
    \includegraphics[width=0.95\linewidth]{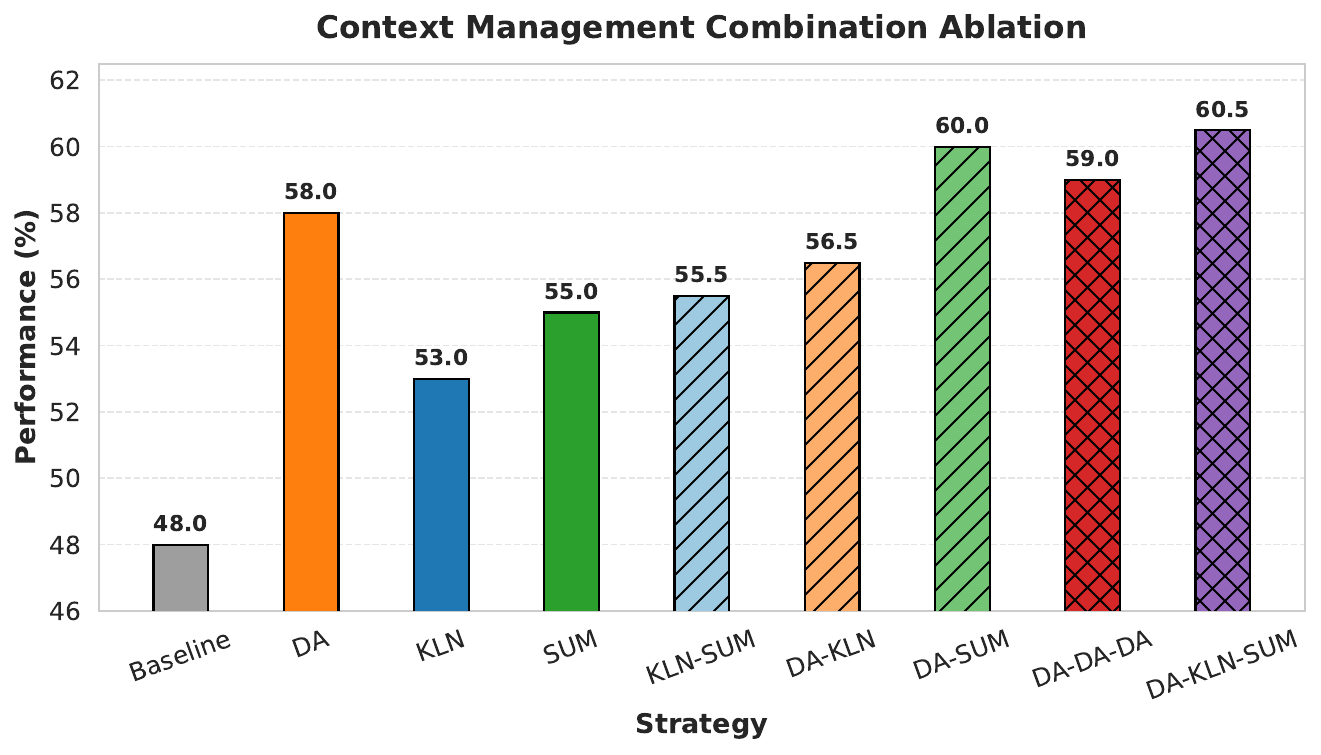}
    \caption{Performance of Tongyi-DR-30B-A3B on BrowseComp under AgentSwing with different context management combinations.}
    \vspace{-1.5em} 
    \label{fig:cm_combination}
\end{wrapfigure}

Figure~\ref{fig:cm_combination} compares different candidate context management combinations within AgentSwing. 
Although some single strategies, especially \textit{Discard-All}, already perform strongly, combining multiple candidate strategies generally yields further gains. This is particularly clear for combinations such as \textit{Discard-All + Summary}, which outperform either constituent strategy used alone. These results suggest that different context management operations provide complementary advantages, and that AgentSwing benefits from routing over a richer set of candidate continuations than any single static policy can offer. More broadly, they indicate that AgentSwing's effectiveness depends not only on the routing mechanism itself, but also on the diversity and complementarity of the candidate strategy set. This also suggests that exploring richer or more specialized candidate strategies is a promising direction for further improving performance.

\section{Analysis of Strategy Transitions under AgentSwing}
\label{app:trans}
\begin{figure*}[ht]
    \centering
    \includegraphics[scale=0.34]{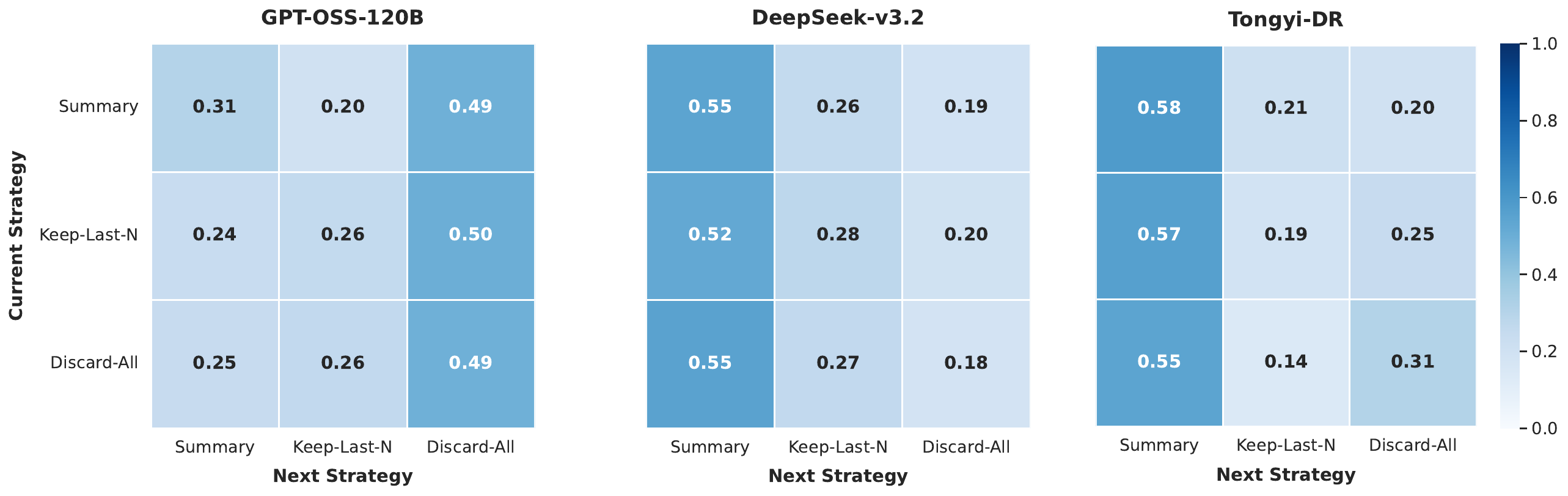}
    \caption{Statistical analysis of strategy transitions under AgentSwing. Each entry gives the probability of routing to the next strategy conditioned on the current one.} 
    \vspace{-3mm}
    \label{fig:strategy_transition}
\end{figure*}

Figure~\ref{fig:strategy_transition} shows the empirical strategy-transition probabilities under AgentSwing. The transition matrices are clearly non-uniform, indicating that routing behavior is not random. Instead, the preferred transitions depend on the underlying backbone. DeepSeek-v3.2 and Tongyi-DR tend to favor \textit{Summary}, whereas GPT-OSS-120B more often transitions to \textit{Discard-All}.

\section{Case Study}
\label{app:case_study}

Tables~\ref{tab:main_branch_mando_part1} and~\ref{tab:main_branch_mando_part2} provide a more detailed turn-by-turn summary of the main trajectory in the "Mando" case from Figure~\ref{fig:case_study}, while Table~\ref{tab:lookahead_branches_mando} reports the corresponding lookahead continuations under different context management strategies.

In addition, Tables~\ref{tab:main_branch_crickets_part1},~\ref{tab:main_branch_crickets_part2}, and~\ref{tab:lookahead_branches_crickets} present a complementary case from GPT-OSS-120B in which \textit{Discard-All} is selected as the context management strategy. This case demonstrates that long-horizon web agents may become trapped in trajectories dominated by substantial noise, repetition, and failed local exploration. In such situations, aggressively resetting the accumulated context through \textit{Discard-All} allows the agent to recover a cleaner search state, refocus on the most relevant clues, and ultimately improve the chance of reaching the correct answer.

\begin{table*}[t]
\centering
\small
\setlength{\tabcolsep}{4pt}
\begin{tabular}{p{1.2cm}p{1.2cm}p{4.4cm}p{7.8cm}}
\toprule
\textbf{Turn ID} & \textbf{Tool Call} & \textbf{Content} & \textbf{Key State} \\
\midrule
Turn 1 & search & 
American rapper born in October in the 1990s; Libra zodiac sign; joined a gang at 14. &
Started with broad candidate retrieval based on birth month, zodiac sign, and gang involvement; surfaced candidates such as Lil Durk. \\
\midrule

Turn 2 & search & 
Rapper joined gang at age 14; American rapper gang member at 14. &
Narrowed the search to rappers who explicitly joined a gang at age 14; exposed promising clues including \$tupid Young and Nipsey Hussle. \\
\midrule

Turn 3 & visit & 
Nipsey Hussle (Wikipedia) &
Verified that Nipsey Hussle was born in August and therefore did not satisfy the October/Libra clue. \\
\midrule

Turn 4 & search & 
October-born rapper who joined a gang at 14; Libra rapper with gang affiliation. &
Searched again for October-born gang-affiliated rappers; Lil Durk became a stronger candidate. \\
\midrule

Turn 5 & visit & 
Lil Durk (Wikipedia) &
Confirmed that Lil Durk was born on October 19, 1992 (Libra), but his gang-entry age remained unsupported and appeared inconsistent with the clue. \\
\midrule

Turn 6 & search & 
Lil Durk joined gang at 14; Durk Banks gang age. &
Tried to verify whether Lil Durk joined a gang at age 14; no direct evidence supported this claim. \\
\midrule

Turn 7 & search & 
Lil Durk viral hit (2015--2020); featured rapper whose father served 15 years. &
Shifted from identity matching to song matching; attempted to find a Lil Durk viral song and a featured rapper with the father-in-prison clue, but results were inconclusive. \\
\midrule

Turn 8 & search & 
Rapper whose father served 15 years in prison. &
Focused on the featured-artist clue; Hit-Boy emerged as the strongest match. \\
\midrule

Turn 9 & visit & 
Hit-Boy (Wikipedia) &
Checked Hit-Boy's profile; found ambiguity because one source suggested 9 years, creating tension with the 15-year clue. \\
\midrule

Turn 10 & search & 
Lil Durk collaborations in 2015--2018. &
Collected candidate Lil Durk collaborations and featured artists for downstream filtering. \\
\midrule

Turn 11 & visit & 
Lil Durk discography (Wikipedia) &
Examined Lil Durk's discography in detail to match songs, years, and featured artists. \\
\midrule

Turn 12 & search & 
Rapper whose father was in prison for 15 years. &
Revisited the father clue; Hit-Boy remained the dominant candidate. \\
\midrule

Turn 13 & visit & 
Hit-Boy's father article (Billboard) &
Confirmed through Billboard that Hit-Boy's father had indeed served 15 years, resolving the earlier inconsistency. \\
\midrule

Turn 14 & search & 
Lil Durk and Hit-Boy collaboration. &
Looked for Lil Durk--Hit-Boy collaborations; found ``Loyal To A Fault,'' but it was released in 2021 and thus fell outside the required window. \\
\midrule

Turn 15 & search & 
Polo G father prison; Lil Baby father prison; King Von father prison. &
Explored alternative featured rappers connected to Lil Durk's songs, but none fit the father clue clearly. \\
\midrule

Turn 16 & search & 
October-born rapper who joined a gang at 14. &
Reopened the identity search because the Lil Durk hypothesis was weakening; results remained noisy. \\
\bottomrule
\end{tabular}
\caption{Main-branch trajectory for the "Mando" case (Part I).}
\label{tab:main_branch_mando_part1}
\end{table*}

\begin{table*}[t]
\centering
\small
\setlength{\tabcolsep}{4pt}
\begin{tabular}{p{1.2cm}p{1.2cm}p{4.4cm}p{7.8cm}}
\toprule
\textbf{Turn ID} & \textbf{Tool Call} & \textbf{Content} & \textbf{Key State} \\
\midrule
Turn 17 & search & 
Hit-Boy father 15-year sentence; Big Hit prison timeline. &
Further validated the 15-year imprisonment timeline around Hit-Boy's father, strengthening the father-side clue. \\
\midrule

Turn 18 & search & 
Songs featuring Hit-Boy in 2015--2018. &
Looked for songs where Hit-Boy was explicitly credited as a featured artist; results were still fragmented. \\
\midrule

Turn 19 & visit & 
Racks in the Middle (Wikipedia) &
Checked ``Racks in the Middle'' and confirmed that although Hit-Boy was involved, Nipsey Hussle did not fit the October/1990s clue. \\
\midrule

Turn 20 & search & 
Viral songs featuring Hit-Boy in 2017--2019. &
Continued searching for a viral song featuring Hit-Boy; no decisive match was found yet. \\
\midrule

Turn 21 & search & 
October-born Crips/Bloods member in the 1990s. &
Explored West Coast gang-affiliated rappers born in October; results were mostly generic gang references. \\
\midrule

Turn 22 & search & 
January 21, 2016 and father served 15 years in prison. &
Investigated the date anchor around January 21, 2016, but this direction produced little value. \\
\midrule

Turn 23 & search & 
I joined a gang at 14; joined the Crips/Bloods at 14. &
Reframed the search around explicit self-reports of joining a gang at 14; this surfaced the critical \textbf{\$tupid Young} clue. \\
\midrule

CM & -- & 
Context management triggered. &
The router selected the \textbf{Keep-Last-N} branch so that it can continue with \$tupid Young clue. \\
\midrule

Turn 24 & visit & 
\$tupid Young (Wikipedia page lookup) &
Attempted to open a Wikipedia page for \textbf{\$tupid Young}, but the page was unavailable. \\
\midrule

Turn 25 & search & 
\$tupid Young birthday; Stupid Young birth date. &
Verified the rapper's identity: \$tupid Young was born on October 3, 1992, matching both the 1990s and Libra clues. \\
\midrule

Turn 26 & search & 
\$tupid Young viral song; \$tupid Young hit song; \$tupid Young featuring. &
Searched his discography and identified ``Mando'' as a strong candidate due to popularity and its feature with Mozzy. \\
\midrule

Turn 27 & search & 
Mozzy father prison 15 years; Mozzy dad penitentiary. &
Verified the featured-artist clue; found a Vice article stating that Mozzy's father had spent 15 years in prison. \\
\midrule

Turn 28 & visit & 
Mando (Wikipedia page lookup) &
Attempted to visit a dedicated page for ``Mando,'' but no useful Wikipedia entry was available. \\
\midrule

Turn 29 & search & 
Mando by \$tupid Young and Mozzy; release date; album information. &
Confirmed that ``Mando'' was released on November 17, 2017 and linked to the 2018 album \textit{One of One}. \\
\midrule

Turn 30 & visit & 
Mando official YouTube video &
Used the YouTube page to confirm the song's viral status via its massive view count (68M+). \\
\midrule

Turn 31 & answer & 
-- &
Integrated all evidence and finalized the answer: \textbf{``Mando''}. \\
\bottomrule
\end{tabular}
\caption{Main-branch trajectory for the "Mando" case (Part II).}
\label{tab:main_branch_mando_part2}
\end{table*}


\begin{table*}[t]
\centering
\small
\setlength{\tabcolsep}{4pt}
\begin{tabular}{p{1.8cm} p{1.5cm} p{1.3cm} p{3.6cm} p{5.4cm}}
\toprule
\textbf{Branch} & \textbf{Turn ID} & \textbf{Tool Call} & \textbf{Content} & \textbf{Key State} \\
\midrule

\space & LA Turn 1 & search &
American rapper born in October in the 1990s; Libra zodiac sign; joined a gang at 14; father served 15 years in prison. &
Restarted from scratch with broad retrieval. It surfaced generic October-born candidates such as Lil Durk, but made no decisive progress on the real target. \\

Discard-All & LA Turn 2 & search &
Rapper whose father served 15 years in prison. &
Focused on the father clue and rediscovered Hit-Boy, but still lacked a correct main-rapper hypothesis. \\

\space & LA Turn 3 & search &
January 21, 2016 and father served 15 years in prison. &
Pursued the date-anchored clue without traction. This branch remained broad and under-focused. \\
\midrule

\space & LA Turn 1 & visit & \$tupid Young (Wikipedia page lookup) &
Attempted to open a Wikipedia page for \$tupid Young, but the page was unavailable. \\

Keep-Last-N & LA Turn 2 & search &
\$tupid Young birthday; Stupid Young birth date. &
Verified the rapper's identity: \$tupid Young was born on October 3, 1992, matching both the 1990s and Libra clues. \\

\space & LA Turn 3 & search &
\$tupid Young viral song; \$tupid Young hit song; \$tupid Young featuring. &
Searched his discography and identified ``Mando'' as a strong candidate due to popularity and its feature with Mozzy. \\
\midrule

\space & LA Turn 1 & search &
Lil Durk joined gang at 14; Lil Durk--Hit-Boy collaborations in 2015--2020. &
Continued from a compressed summary centered on Lil Durk and Hit-Boy. This preserved structure but also inherited a misleading focus. \\

Summary & LA Turn 2 & visit &
Lil Durk (Wikipedia); Lil Durk discography (Wikipedia). &
Verified that Lil Durk joined the Black Disciples at age 17 rather than 14, and found no Lil Durk--Hit-Boy collaboration within 2015--2020. \\

\space & LA Turn 3 & search &
October-born Libra rapper who joined a gang at 14. &
Only after falsifying the Lil Durk path did this branch begin searching for alternative rappers; within the lookahead horizon, it did not reach the \$tupid Young breakthrough. \\
\bottomrule
\end{tabular}
\caption{Lookahead branches triggered by context management in the "Mando" case.}
\label{tab:lookahead_branches_mando}
\end{table*}

\begin{table*}[t]
\centering
\small
\setlength{\tabcolsep}{4pt}
\begin{tabular}{p{1.2cm} p{1.2cm} p{5cm} p{7.8cm}}
\toprule
\textbf{Turn ID} & \textbf{Tool Call} & \textbf{Content} & \textbf{Key State} \\
\midrule

Turn 1 & search &
Performer who stapled paper to his forehead; sideshow stapling act. &
Started with direct retrieval on the stapling clue, but results were dominated by noisy modern pages, social media posts, and irrelevant literal uses of ``stapled paper.'' \\
\midrule

Turn 2 & search &
Paper-to-forehead sideshow performer who ate something live. &
Tried to combine the stapling clue with the ``ate live creatures'' clue; surfaced sideshow-related entities such as Jim Rose Circus, but no stable performer identity. \\
\midrule

Turn 3 & search &
Strongwoman associated with ``beef, game, and plenty of vegetables.'' &
Shifted to a secondary clue in the question, but the retrieved results were largely noisy and did not yet identify the relevant strongwoman. \\
\midrule

Turn 4 & search &
Bethel, Connecticut; Manhattan museum; Feejee Mermaid. &
Used the Bethel / museum / Feejee Mermaid clue to infer the publication domain; this pointed toward P.\ T.\ Barnum and the historical oddities / sideshow space. \\
\midrule

Turn 5 & search &
Bethel, Connecticut; sideshow; Manhattan museum; Feejee Mermaid. &
Repeated the supporting-entity search, but the results were still not sufficiently specific to identify the source publication. \\
\midrule

Turn 6 & search &
Strongwoman who threw a heckler across a tent. &
Switched to another distinctive supporting clue in order to identify the common source through a secondary figure. \\
\midrule

Turn 7 & search &
Strongwoman threw a heckler. &
A simplified version of the query surfaced references to Minerva, helping move the search toward historical strongwoman material. \\
\midrule

Turn 8 & visit &
Victorian strongwomen article (iNews). &
Visited the article and confirmed that the strongwoman was Josephine Schauer Blatt (Minerva), establishing that the question belongs to the historical sideshow / freak-show domain. \\
\midrule

Turn 9 & search &
Circus performer who staples paper to his forehead. &
Returned to the stapling clue after confirming the domain; the results now included more circus / sideshow-related pages, but still no exact match. \\
\midrule

Turn 10 & visit &
Jelly Boy the Clown article (East Bay Times). &
Found a modern performer who allowed money to be stapled to his face, but this did not match the clue about eating live creatures. \\
\midrule

Turn 11 & search &
Stapling performer who also eats live creatures. &
Tried to jointly resolve the two key attributes, but the results still lacked a decisive source text. \\
\midrule

Turn 12 & search &
Paper on forehead; eating live creatures. &
Continued direct clue search, but the retrieval remained noisy and failed to identify the exact publication or performer. \\
\midrule

Turn 13 & search &
Exact phrase: ``stapled paper to his forehead.'' &
Achieved the first major breakthrough: search results surfaced the \textit{PDF} \textbf{The Victorian Sideshow}, with a snippet containing the critical phrase ``has willingly stapled paper to his forehead ... eaten a mouthful ...'' \\
\midrule

Turn 14 & visit &
The Victorian Sideshow PDF (direct access attempt). &
Tried to open the PDF directly, but the tool returned no extractable content. This established the central bottleneck of the case. \\
\midrule

Turn 15 & search &
The Victorian Sideshow PDF. &
Searched for alternative paths to the same PDF, but the results still pointed back to the same inaccessible source. \\
\midrule

Turn 16 & visit &
The Victorian Sideshow PDF (second access attempt). &
Repeated the PDF visit attempt, but the extraction failure persisted. \\
\midrule

Turn 17 & search &
Paper to his forehead; sideshow. &
Looked for alternative source surfaces after the failed PDF access; results still pointed mainly to the same PDF and its mirrors. \\
\midrule

Turn 18 & visit &
Scribd mirror of \textit{Sideshow}. &
Attempted to recover the content through Scribd, but the page was effectively inaccessible. \\

\bottomrule
\end{tabular}
\caption{Main-branch trajectory for the "live-crickets" case (Part I).}
\label{tab:main_branch_crickets_part1}
\end{table*}

\begin{table*}[t]
\centering
\small
\setlength{\tabcolsep}{4pt}
\begin{tabular}{p{1.2cm} p{1.2cm} p{5.6cm} p{7.2cm}}
\toprule
\textbf{Turn ID} & \textbf{Tool Call} & \textbf{Content} & \textbf{Key State} \\
\midrule

Turn 19 & search &
Stapled paper; forehead; sideshow; eaten. &
Combined the snippet clues again, but the results still revolved around the unresolved PDF source. \\

\midrule
Turn 20 & search &
Full snippet phrase including ``eaten a mouthful.'' &
Queried the visible snippet directly; this helped confirm the source phrase, but still did not reveal the missing object after ``eaten a mouthful of ...'' \\

\midrule

Turn 21 & search &
Performer name from the stapling clue. &
Tried to infer the performer identity directly from the snippet description, but the retrieval remained inconclusive. \\
\midrule

Turn 22 & search &
Exact phrase: ``has willingly stapled paper to his forehead.'' &
Repeated exact-phrase retrieval to localize the passage more precisely, but still without extractable full text. \\
\midrule

Turn 23 & search &
The Victorian Sideshow PDF. &
Re-confirmed that \textbf{The Victorian Sideshow} was the likely shared source behind the unusual individuals in the question. \\
\midrule

Turn 24 & visit &
The Victorian Sideshow PDF (targeted extraction attempt). &
Made a more targeted attempt to extract the paragraph about the stapling performer and the live-creature clue, but the visit tool still failed. \\
\midrule

CM & -- &
Context management triggered. &
Context management was triggered because the trajectory had become long, noisy, and partially repetitive. The router evaluated three branches and selected \textbf{Discard-All}. \\
\midrule

Turn 25 & search &
Stapled paper to forehead performer. &
After the reset, restarted with a cleaner search plan; quickly re-entered the correct search space without carrying over the accumulated local noise. \\
\midrule

Turn 26 & search &
Feejee Mermaid; Minerva; Jo-Jo; supporting clue bundle. &
Used multiple supporting entities together to verify that the publication family was correct and that the search was grounded in the historical sideshow domain. \\
\midrule

Turn 27 & visit &
Jo-Jo the Dog-Faced Boy article. &
Confirmed another supporting figure from the same source family, increasing confidence that the publication hypothesis was correct. \\
\midrule

Turn 28 & search &
Exact stapling-performer phrasing. &
Returned to the core unresolved clue after re-confirming the correct publication \textbf{The Victorian Sideshow}. \\
\midrule

Turn 29 & visit &
The Victorian Sideshow PDF. &
Direct extraction still failed, confirming that the bottleneck was tool-access related rather than search-related. \\
\midrule

Turn 30 & visit &
Alternative text-extraction endpoint for the PDF. &
Achieved the decisive breakthrough by using an alternative access path that successfully returned the source text, revealing that the performer had ``eaten a mouthful of live crickets.'' \\
\midrule

Turn 31 & answer &
-- &
Integrated all evidence and produced the final answer: the person who stapled paper to his forehead ate a mouthful of \textbf{live crickets}. \\
\bottomrule
\end{tabular}
\caption{Main-branch trajectory for the "live-crickets" case (Part II).}
\label{tab:main_branch_crickets_part2}
\end{table*}

\begin{table*}[t]
\centering
\small
\setlength{\tabcolsep}{4pt}
\begin{tabular}{p{1.8cm} p{1.5cm} p{1.3cm} p{3.8cm} p{5.6cm}}
\toprule
\textbf{Branch} & \textbf{Turn ID} & \textbf{Tool Call} & \textbf{Content} & \textbf{Key State} \\
\midrule

\space & LA Turn 1 & search &
Stapled paper to forehead performer. &
Restarted from scratch and quickly re-entered the correct search space around the stapling-performer clue, without inheriting the noisy local loop around failed PDF extraction. \\

Discard-All & LA Turn 2 & search &
Feejee Mermaid; Minerva; Jo-Jo; supporting clue bundle. &
Used multiple supporting entities together to verify that the publication family was correct and that the search was grounded in the historical sideshow domain. \\

\space & LA Turn 3 & visit &
Jo-Jo the Dog-Faced Boy article. &
Also revisited supporting clues from the question, indicating that the branch was reconstructing the source-publication hypothesis through multiple entities rather than overfitting to one failed access path. \\
\midrule

\space & LA Turn 1 & visit &
Scribd mirror of \textit{Sideshow}. &
Preserved the most recent local context, which was already dominated by failed source-extraction attempts; immediately re-entered the same bottleneck. \\

Keep-Last-N & LA Turn 2 & search &
The Victorian Sideshow PDF. &
Continued searching for alternative access points to the same PDF, but remained trapped in the same unresolved extraction problem. \\

\space & LA Turn 3 & visit &
The Victorian Sideshow PDF. &
Attempted direct PDF access again and failed, showing that preserving the most recent context mainly preserved the local dead end rather than useful progress. \\
\midrule

\space & LA Turn 1 & search &
Exact stapling phrase; sideshow; live-creature clue. &
Used the summary-preserved hypothesis that The Victorian Sideshow was likely the correct source, and re-centered search on the key unresolved phrase. \\

Summary & LA Turn 2 & search &
Repeated phrase-centered retrieval. &
Continued operating at the correct abstraction level, but still remained dependent on search-result snippets and inaccessible source pages. \\

\space & LA Turn 3 & search &
Repeated snippet-oriented search behavior. &
Maintained a cleaner high-level focus than Keep-Last-N, but did not produce a concrete recovery step that would break the extraction bottleneck. \\
\bottomrule
\end{tabular}
\caption{Lookahead branches triggered by context management in the "live-crickets" case.}
\label{tab:lookahead_branches_crickets}
\end{table*}


\end{document}